\documentclass[10pt,twocolumn,letterpaper]{article}

\usepackage{cvpr}              

\usepackage{graphicx}
\usepackage{amsmath}
\usepackage{amssymb}
\usepackage{booktabs}

\usepackage[pagebackref,breaklinks,colorlinks]{hyperref}
\usepackage{color}

\usepackage[capitalize]{cleveref}
\crefname{section}{Sec.}{Secs.}
\Crefname{section}{Section}{Sections}
\Crefname{table}{Table}{Tables}
\crefname{table}{Tab.}{Tabs.}

\def\cvprPaperID{*****} 
\def\confName{CVPR}
\def\confYear{2022}

\begin{document}

\title{
  OMG: Observe Multiple Granularities for Natural Language-Based Vehicle Retrieval
}

\author{Yunhao Du$^{1}$\footnotemark[1]\ , Binyu Zhang$^{1}$\footnotemark[1]\ , Xiangning Ruan$^1$, Fei Su$^{1,2}$, Zhicheng Zhao$^{1,2}$\footnotemark[2], Hong Chen$^3$\\
$^1$Beijing University of Posts and Telecommunications \\
$^2$Beijing Key Laboratory of Network System and Network Culture, China\\
$^3$China Mobile Research Institute \\
{\tt\small \{dyh\_bupt,zhangbinyu,ruan,zhaozc,sufei\}@bupt.edu.cn} \\
{\tt\small chenhongyj@chinamobile.com}
}
\maketitle
\renewcommand{\thefootnote}{\fnsymbol{footnote}}
\footnotetext[1]{Equal contribution}
\footnotetext[2]{Corresponding authors}

\begin{abstract}
   Retrieving tracked-vehicles by natural language descriptions plays a critical role in smart city construction.
   It aims to find the best match for the given texts from a set of tracked vehicles in surveillance videos.
   Existing works generally solve it by a dual-stream framework, which consists of a text encoder, a visual encoder and a cross-modal loss function.
   Although some progress has been made, they failed to fully exploit the information at various levels of granularity.
   To tackle this issue, we propose a novel framework for the natural language-based vehicle retrieval task, \textbf{OMG}, 
   which \textbf{O}bserves \textbf{M}ultiple \textbf{G}ranularities with respect to visual representation, textual representation and objective functions.
   For the visual representation, target features, context features and motion features are encoded separately.
   For the textual representation, one global embedding, three local embeddings and a color-type prompt embedding 
   are extracted to represent various granularities of semantic features.
   Finally, the overall framework is optimized by a cross-modal multi-granularity contrastive loss function.
   Experiments demonstrate the effectiveness of our method.
   Our OMG significantly outperforms all previous methods and ranks the 9th on the 6th AI City Challenge Track2.
   The codes are available at \href{https://github.com/dyhBUPT/OMG}{https://github.com/dyhBUPT/OMG}.
\end{abstract}

\section{Introduction}
\label{sec:intro}

\begin{figure}[t]
  \centering
  \includegraphics[width = 0.45\textwidth]{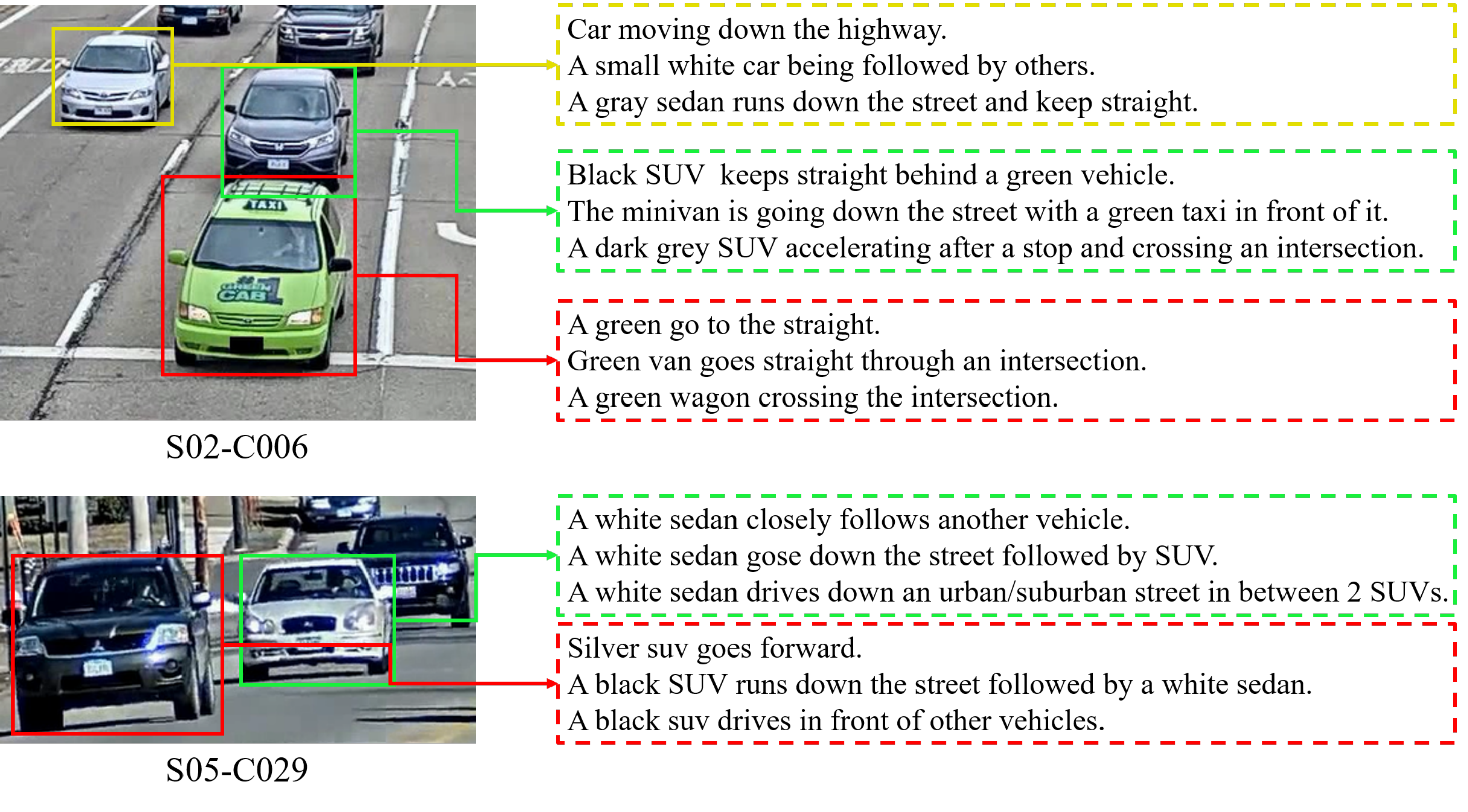}
  \caption{
    Visualization examples of the natural language-based vehicle retrieval task.
    The query generally describes the color, type, motion state and the relations (with neighbor objects) of the target vehicle.
  }
  \label{figure_task}
\end{figure}

With the development of the artificial intelligence technology, smart transportation and city has elicited increasing attention in recent years.
Vehicle retrieval, which aims to retrieve tracked vehicles given a query, occupies an important position.
Most previous works solve this problem under a purely visual setting, 
which try to identify the images or tracklets of the same vehicle across different surveillance camera views \cite{liu2016deep,liu2016deep2,tang2019cityflow}.
However, image-based setting limits its application scenarios, bacause an image query is not always available.
Comparatively speaking, Natural Language (NL) descriptions can be one of the most convenient way to give a query \cite{feng2021cityflow}.
Therefore, it is not trivial to develop an effective system for natural language-based vehicle retrieval.
Figure \ref{figure_task} gives example frames, bounding boxes and corresponding descriptions from the CityFlow-NL dataset \cite{feng2021cityflow}, 
which is a large-scale benchmark for the text-based vehicle retrieval task.

Existing works tend to solve this task by using a dual-stream framework \cite{ging2020coot} because of its flexibility and efficiency \cite{liu2021hit}.
This framework consists of separate visual encoder and text encoder to extract modal-specific representations respectively and then computes their similarity.
Then it is optimized by InfoNCE loss \cite{van2018representation}, Instance loss \cite{zheng2020dual}, etc.
Though these methods have made some progress, they fail to mine adequate multi-granularity information.
For instance, the relationships between the target vehicle and its neighbors is ignored in \cite{1st, 2nd, 7th, 8th},
and \cite{4th, 11th} fail to fully utilze the color and type information. 
Moreover, they don't take good advantage of rich textual granular features.

In this paper, we propose OMG, which observes multiple granularities in an unified framework.
It follows the dual-stream architecture, that is, one visual stream and one textual stream.
The visual stream takes a cropped vehicle image, a context image and a foreground motion map as input, and then extracts their features by three individual image encoders.
As for the textual stream, one global text, three local texts and one prompt are encoded separately by five tied text encoders. 
While training, a multi-granularity InfoNCE loss is applied to perform cross-modal contrastive learning between them.
Furthermore, an ID loss is also used for the visual stream for more discriminative representation.

Extensive experiments verify the effectiveness of our proposed methods.
To be specific, by taking "ResNet50 + BERT + InfoNCE" as baseline \cite{he2016deep, turc2019well, van2018representation},
a series of proposed optimization strategies boost its performance by \textbf{144\%} on the validation set (MRR from 0.208 to 0.507).
Entries based on the presented methodology ranks the 9th on the AI CIty Challenge 2022 Track2.
The contributions of this work are summarized as follows:

1) We propose a novel framework for the natural language-based vehicle retrieval task, which fully exploits the multi-granularity information in a joint manner.

2) Various optimization tricks are employed to further improve the representation capability of the network, 
i.e., auxiliary ID loss, backtranslation-based text augmentation, stronger image and text encoders \cite{liu2021swin, radford2021learning}.

3) Extensive experiments prove the effectiveness of our method.

\begin{figure*}[t]
  \centering
  \includegraphics[width = 0.9\textwidth]{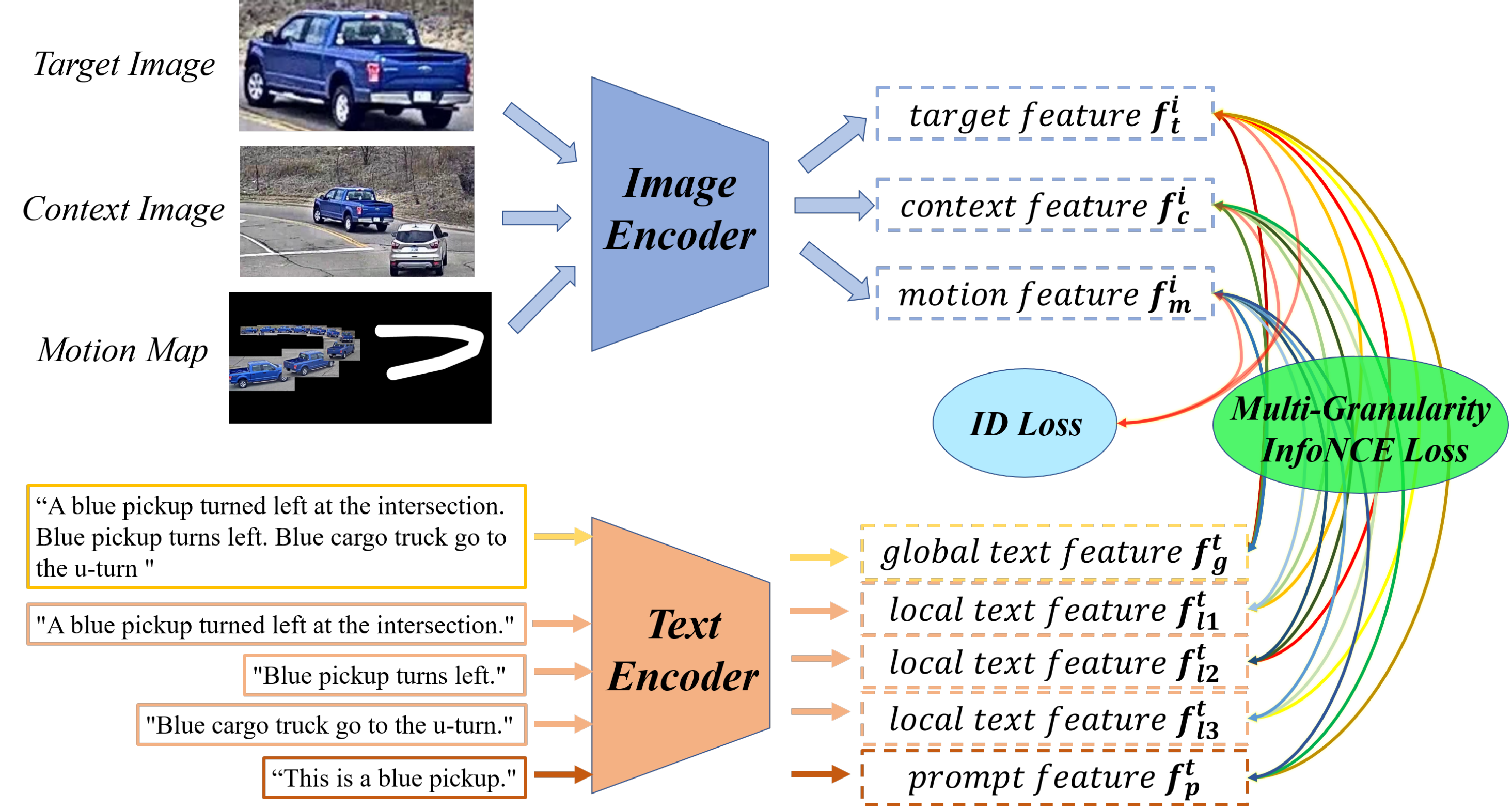}
  \caption{
    Framework of the proposed OMG, which consists of a visual stream and a textual stream.
    For the visual stream, it takes a cropped target image, a context image and a 4-channel foreground motion map as input, 
    and encodes them into a target feature $f^i_t$, a context feature $f^i_c$ and a motion feature $f^i_m$ respectively. 
    As to the textual stream, it embdes one global feature $f^t_g$, three local features $f^t_l$ and one prompt feature $f^t_p$ 
    from different granularity of textual inputs.
    The overall framework is optimized with a multi-granularity InfoNCE loss and an ID loss in an end-to-end manner.
  }
  \label{figure_framework}
\end{figure*}

\section{Related Works}

\subsection{Text-based Video Retrieval}

Text-based video retrieval aims to search the corresponding video via the given text description among a number of videos. 
Due to its great range of potential applications, it has received wide attention by the researchers. The most of existing methods 
adopt contrastive learning to train a network which can estimate the similarity between the video and the text descriptions. 
They take both video and text descriptions as input, and then extract the video features and text features via the modal encoders. 
They tend to update the networks by minimizing the distance between each pair of the video-text features.
Due to the difference of the modality and limitation of the network, the early methods \cite{mithun2017cmu, mithun2018learning, li2018renmin} 
usually use a dual-stream framework, which can reduce the computational complexity. 
These works encode the video by the visual feature extractor (C3D\cite{tran2015learning}, I3D \cite{carreira2017quo}, etc.) and encode the text 
by the textual feature extractor (LSTM \cite{hochreiter1997long}, GRU \cite{cho2014learning}, etc.), 
and then estimate the video-text similarity in a potential semantic space.

Recently, with the successful migration of Transformer \cite{vaswani2017attention} from natural language processing to computer vision, 
(ViT \cite{dosovitskiy2020image}, Swin-Transformer \cite{liu2021swin}, etc.), the mainstream methods 
\cite{lu2019vilbert,li2020hero,li2021align,liu2021hit,fang2021clip2video,radford2021learning,hao2021multi,yang2021taco,wu2021hanet,liu2022featinter} on 
video-language retrieval tasks begin to utilize Transformer as encoders for both of the video and the natural language. 
HERO \cite{li2020hero} and ALPRO \cite{li2021align} explore to boost the video-text alignment via the large-scale pre-training tasks. 
HiT \cite{liu2021hit} further proposed a hierarchical model with momentum contrast for video-text retrieval. 
Also, some single-stream networks come out, due to the consistency of the Transformer encoders structure.
ViLBERT \cite{lu2019vilbert} firstly explore to utilize a single-stream Transformer network to retrieve the video based on natural language descriptions, 
which also attempts to align the corresponding video and text features in the potential semantic space. 
Further, due to the success of CLIP \cite{radford2021learning}, which demonstrates the powerful performance of image-text contrastive learning, 
many researchers conduct experiments on the basis of CLIP. 
Besides, CLIP2Video \cite{fang2021clip2video} extends CLIP from image-text retrieval to video-text retrieval. 
Moreover, MFGATN \cite{hao2021multi}, TACo \cite{yang2021taco}, HANet \cite{wu2021hanet} and FeatInter \cite{liu2022featinter} 
also achieve good performance in the video-text retrieval tasks by designing different modules to enhance the comprehension of the video-text pair.

\subsection{Text-based Vehicle Retrieval}

AYCE \cite{11th} proposes a modular solution which applies BERT \cite{turc2019well} to embed textual descriptions 
and a CNN \cite{he2016deep} with a Transformer model \cite{vaswani2017attention} to embed visual information.
SBNet \cite{10th} presents a substitution module that helps project features from different domains into the same space,
and a future prediction module to learn temporal information by predicting the next frame.
Pirazh et al. \cite{8th} and Tam et al. \cite{4th} adopts CLIP \cite{radford2021learning} to extract frame features and textual features.
TIED \cite{7th} proposes an encoder-decoder based model in which the encoder embeds two modalities into the common space 
and the decoder jointly optimizes these embeddings by an input-token-reconstrucion task.
Tien-Phat et al. \cite{6th} adapts COOT \cite{ging2020coot} to model the cross-modal relationships with both appearance and motion attributes.
Eun-Ju et al. \cite{3rd} propose to perform color and type classification for both target and front-rear vehicles, 
and conduct movement analysis based on the Kalman filter algorithm \cite{kalman1960new}.
DUN \cite{2nd} uses pretrained CNN and GloVe \cite{pennington2014glove} to extract modal-specific features 
and GRUs \cite{cho2014properties} to exploit temporal information.
CLV \cite{1st} proposes the simple and effective global motion image, which is jointly encoded with the cropped image and language description,
and ranks first on the AI City Challenge 2021 Track5.

Though these methods have achieved promising results on the CityFlow-NL benchmark, they don't utilze omni-features adequately.
Differently, our method can mine rich information from multiple granularities, which helps extract more discriminative features.

\section{Method}

In this section, we present the details of the OMG framework as shown in Figure \ref{figure_framework}.
Specifically, we start with the visual stream in Section \ref{s1}, 
and then introduce the textual stream in Section \ref{s2}.
Finally, Section \ref{s3} illustrates the multi-granularity InfoNCE loss and the auxiliary ID loss.

\subsection{Multi-Granularity Visual Stream}\label{s1}

For the visual stream, three granularities of images are inputed into image encoders to extract features respectively.
Note that these three encoders don't share weights.

\noindent \textbf{Target Features.} Given a trajectory of the target vehicle, a frame is sampled and the target image is cropped according to the bounding box.
Then it is taken as the input to generate the target feature $f^i_t$.
By setting a high resolution, the network can focus on the details and implicitly encode the appearance information, e.g., color and type.

\noindent \textbf{Context Features.} The target feature only contains the target vehicle's own information by excluding the background, 
ignoring the relationship between neighbors.
Instead, we extend the original bounding box $[x,y,w,h]$ to $[x-w, y-h, 3w, 3h]$ to crop the context image around the target vehicle.
It has a larger receptive field and can usually include nearby vehicles, which helps the encoder embed the relation information and predict the context feature $f^i_c$.

\noindent \textbf{Motion Features.} Another vital information is the motion pattern of the target vehicle.
Inspired by the background modeling based motion image in \cite{1st}, we propose a 4-channel foreground motion map,
which consists of a 3-channel foreground cropped image and a single channel motion image.
For the former one, the whole trajectory of the target vehicle is cropped and pasted into a black image.
Note that we only keep the bounding boxes that don't overlap too much.
For the latter one, the center positions of these bounding boxes are used to form a thick trajectory line.
Finally, these two images are concatenated along the channel dimension and taken as the input of the image encoder.
It's encoded into the motion feature $f^i_m$, which contains both coarse appearance and motion information simutaneously.

\begin{figure*}[t]
  \centering
  \includegraphics[width = 0.9\textwidth]{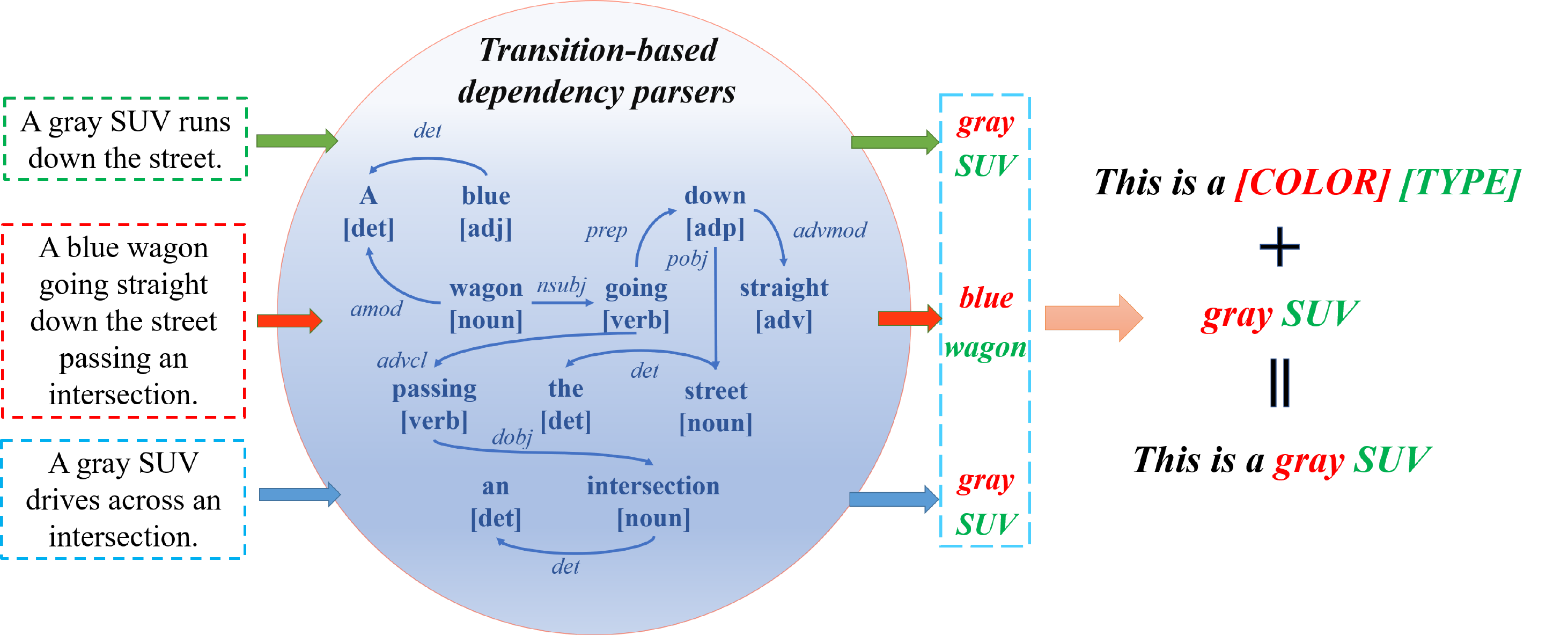}
  \caption{
    The illustration of the color-type prompt generation process,
    which consists of three steps, i.e., dependency parsing, voting and prompt generation.
    Note that we only visualize the dependency parsing results of one sentence for simplicity, 
    i.e., \textit{"A blue wagon going straight down the street passing an intersection"}.
  }
  \label{figure_prompt}
\end{figure*}

\subsection{Multi-Granularity Textual Stream}\label{s2}

Similar with the visual stream, the textual stream also contains three granularities of inputs, i.e., the global text, the local texts and the color-type prompt.
The notable difference between these two streams is that the weights of all text encoders are shared and freezed except for the last fully connected layer.

\noindent \textbf{Global Features.} Every trajectory corresponds to three unique natural language sentences.
The descriptions of these sentences would have different emphases.
For example, some may focus on the appearance and motion information (e.g., \textit{A black sedan makes a right turn at the intersection.}),
and others may contain the relation information (e.g., \textit{A white SUV switches to the left lane followed by another white vehicle.}).
In order to comprehensively consider them, an intuitive and simple way is to concatenate these three sentences into an unified one, termed the global text.
The extracted global feature $f^t_g$ from it includes rich semantic information from a global view.

\noindent \textbf{Local Features.} Though the global feature embeds the integrated semantics, it is too coarse to focus on details.
To make the network pay attention to more fine-grained information, we also input the three sentences into the network respectively 
to predict three local features $f^t_{l1}, f^t_{l2}, f^t_{l3}$.
Different from $f^t_g$, the $f^t_{l*}$ places more emphasis on some specific features, e.g., appearance, motion and relation.

\noindent \textbf{Prompt Features.} The appearance would be the most vital information to distinguish different vehicles.
Inspired by the prompt learning in Natural Language Processing (NLP) \cite{liu2021pre}, we design a color-type prompt for better appearance representation.
The prompt generation process is illustrated in the Figure \ref{figure_prompt}.
Specifically, the color and type of the vehicle are extracted from each sentence first by using a dependency parser \cite{honnibal2015improved}.
Then the final color and type are determined by a simple voting mechanism ("gray" and "SUV" in the figure).
Finally, the color-type prompt is generated by the template \textit{"This is a [COLOR] [TYPE]"}, e.g., \textit{"This is a gray SUV"} in this example.
By using this simple and intuitive prompt as input and predicting the prompt feature $f^t_p$, it is easier to focus on the appearance information for the text encoder.

\begin{table*}[t]
  \begin{center}
    \caption{
      Ablation study of OMG on the validation set.
      “↑” means higher is better.
    }
    \label{table_ablation1}
    \resizebox{0.95\textwidth}{!}{
      \begin{tabular}{cl|c|c|c|c|c|c|c|c|c|c|c|c|c}
        \toprule[1pt]
        & \textbf{Method} & \textbf{BoT} & \textbf{Context} & \textbf{Motion} & \textbf{Local} & \textbf{Prompt} & \textbf{CLIP} 
        & \textbf{NLAug} & \textbf{IDLoss} & \textbf{Swin} & \textbf{MFT} & \textbf{MRR(↑)} & \textbf{R@5(↑)} & \textbf{R@10(↑)}\\
        \hline
        & Baseline &  &  &  &  &  &  &  &  &  &  & 0.208 & 0.269 & 0.363 \\ 
        &  & \checkmark &  &  &  &  &  &  &  &  &  & 0.218 & 0.285 & 0.382 \\ 
        &  & \checkmark & \checkmark &  &  &  &  &  &  &  &  & 0.228 & 0.315 & 0.422 \\ 
        &  & \checkmark & \checkmark & \checkmark &  &  &  &  &  &  &  & 0.204 & 0.263 & 0.435 \\ 
        &  & \checkmark &  &  & \checkmark &  &  &  &  &  &  & 0.241 & 0.333 & 0.457 \\ 
        &  & \checkmark &  &  & \checkmark & \checkmark &  &  &  &  &  & 0.273 & 0.384 & 0.505 \\ 
        &  & \checkmark & \checkmark &  & \checkmark & \checkmark &  &  &  &  &  & 0.285 & 0.411 & 0.556 \\ 
        &  & \checkmark & \checkmark &  & \checkmark & \checkmark & \checkmark &  &  &  &  & 0.316 & 0.435 & 0.586 \\ 
        &  & \checkmark & \checkmark &  & \checkmark & \checkmark & \checkmark & \checkmark &  &  &  & 0.401 & 0.567 & 0.723 \\ 
        &  & \checkmark & \checkmark &  & \checkmark & \checkmark & \checkmark & \checkmark & \checkmark &  &  & 0.438 & 0.591 & 0.750 \\ 
        &  &  & \checkmark &  & \checkmark & \checkmark & \checkmark & \checkmark & \checkmark & \checkmark &  & 0.490 & \textbf{0.672} & 0.790 \\
        & OMG &  & \checkmark &  & \checkmark & \checkmark & \checkmark & \checkmark & \checkmark & \checkmark & \checkmark & \textbf{0.507} & 0.667 & \textbf{0.812} \\
        \bottomrule[1pt]
      \end{tabular}
    }
  \end{center}
\end{table*}

\subsection{Multi-Granularity Loss}\label{s3}

InfoNCE loss \cite{van2018representation} is a widely used objective function for contrastive learning, which is then extended to cross-modal tasks.
In this work, we apply it to our cross-modal multi-granularity framework and name it \textit{Multi-Granularity InfoNCE loss}.
Given a batch with $M$ text-vehicle pairs $\{(t_i, v_i)\}_{i=1}^{M}$, we first extract their multi-granularity features.
Particularly, every $t_i$ is embedded into $N_t$ features $\{t_i^j\}_{j=1}^{N_t}$ and every $v_i$ is embedded into $N_v$ features $\{v_i^k\}_{k=1}^{N_v}$,
where $N_t=5$ and $N_v=3$ in this work.
Then the multi-granularity InfoNCE loss from text to image is:
\begin{equation}
  L_{info}^{t2i} = {1 \over {M N_t N_v}} \sum^M_{i=1} \sum^{N_t}_{j=1} \sum^{N_v}_{k=1} 
  log({{e^{s(t^j_i, v^k_i) / \tau}} \over {\sum^M_{n=1}e^{s(t^j_i, v^k_n) / \tau}}}), \label{a}
\end{equation}
\noindent where $\tau$ denotes a learnable temperature parameter and $s(\cdot, \cdot)$ measures the cosine similarity as:
\begin{equation}
  s(u, v) = {{u^Tv} \over {||u|| \ ||v||}}. \label{b}
\end{equation}
\noindent Similarly, the image-to-text one is: 
\begin{equation}
  L_{info}^{i2t} = {1 \over {M N_t N_v}} \sum^M_{i=1} \sum^{N_t}_{j=1} \sum^{N_v}_{k=1} 
  log({{e^{s(v^k_i, t^j_i) / \tau}} \over {\sum^M_{n=1}e^{s(v^k_i, t^j_n) / \tau}}}). \label{c}
\end{equation}
\noindent Finally, the whole multi-granularity InfoNCE loss is:
\begin{equation}
  L_{info} = (L_{info}^{t2i} + L_{info}^{i2t}) / 2. \label{d}
\end{equation}

Though $L_{info}$ can help improve the representational ability in a contrastive learning manner, 
it treats all samples as the negatives except that in the anchor pair.
However, some vehicles from different text-vehicle pairs have the same ID.
To be specific, there are 2,155 pairs in the traning set of the CityFlow-NL dataset \cite{feng2021cityflow}, but they are from $C=482$ vehicles from different views.
It would deteriorate the feature learning by treating these vehicles with the smae ID as negative samples, especially for the image encoders.
For that reason, we introduce the cross-entropy loss on the multi-granularity visual features $\{v_i^k\}_{k=1}^{N_v}$ as an auxiliary ID loss.
Given a vehicle  $v_i$, we denote $y_i$ as the truth ID label and $p_{i,c}^k$ as the ID prediction logits of class $c$ for granularity $k$.
The ID loss is computed as follows:
\begin{equation}
  L_{id} = {1 \over {M N_v}} \sum^M_{i=1} \sum^C_{c=1} \sum^{N_v}_{k=1} -q_c log(p_{i,c}^k), \label{e}
\end{equation}
\noindent where $q_c = 1$ if $y_i=c$, otherwise $q_c = 0$.

Finally, the overall objective function is formulated as:
\begin{equation}
  L_{OMG} = \lambda_1 L_{info} + \lambda_2 L_{id}, \label{f}
\end{equation}
\noindent where $\lambda_1, \lambda_2$ are the weights for the two losses. 
We set $\lambda_1 = 1, \lambda_2 = 1$ in our experiments.

\section{Experiments}

\subsection{Dataset and Evaluation}

We use the CityFlow-NL dataset \cite{feng2021cityflow} to train and evaluate our model, 
and it is extended from the CityFlow benchmark \cite{tang2019cityflow} by annotating natural language descriptions for vehicle targets.
It is the first city-scale multi-target multi-camera tracking with natural language descriptions dataset
that provides precise descriptions for multi-view ground truth vehicle tracks.
The CityFlow-NL dataset consists of 666 targets vehicles in 3,028 single-view tracks and 5,289 unique NL descriptions.
For the AI City Challenge, the dataset contains 2,498 tracks of vehicles with three descriptions.
530 unique vehicle tracks together with 530 query sets each with three descriptions are curated for this challenge. 

The natural language-based vehicle retrieval task uses the Mean Reciprocal Rank (MRR) as the main evaluation metric.
Specifically, the reciprocal rank of a query response is the multiplicative inverse of the rank of the first correct answer.
The mean reciprocal rank is the average of the reciprocal ranks of the overall test set:
\begin{equation}
  MRR = {1 \over N} \sum^N_{i=1} {1 \over {rank_i}}, \label{g}
\end{equation}
where $rank_i$ is the rank position of the first correct answer for the $i$-th query.
Besides, Recall@5 and Recall@10 are also evaluated.

\subsection{Implement Details}

In the training phase, we train OMG with batch size 24 on three 16G Tesla T4 GPUs. 
For the visual stream, we first randomly sample a frame from the target trajectory.
Then the input images are resized to 384 $\times$ 384.
\textit{RandomCrop} and \textit{RandomApply} are used to perform data augmentation.
For the textual stream, the backbone is freezed except the last fully connected layer.
The cosine annealing strategy is used with a base learning rate 6.7e-3 and a minimum learning rate 6.7e-6.
The network is trained with the AdamW \cite{loshchilov2017decoupled} optimizer for 600 epochs with 300 warm-up epochs.

While inference, the middle frame of the trajectory is sampled as the input of the visual stream.
All $N_t N_v$ pairs of cross-modal multi-granularity of features are used to compute pairwise similarities.
Then they are simply averaged for the final similarity prediction.

\begin{table*}[t]
  \begin{center}
    \caption{
      Ablation study of the OSG on the validation set. "↑" means higher is better.
    }
    \label{table_ablation2}
    \resizebox{0.75\textwidth}{!}{
      \begin{tabular}{cl|c|c|c|c|c|c|c|c|c|c}
        \toprule[1pt]
        & \textbf{Method} & \textbf{NLAug} & \textbf{Color} & \textbf{Type} & \textbf{CLIP} & \textbf{ID} & \textbf{Motion} & \textbf{Swin} 
        & \textbf{MRR(↑)} & \textbf{R@5(↑)} & \textbf{R@10(↑)}\\
        \hline
        & Baseline &            &            &            &            &            &            &            & 0.208 & 0.269 & 0.363 \\ 
        &          & \checkmark &            &            &            &            &            &            & 0.212 & 0.293 & 0.401 \\ 
        &          & \checkmark & \checkmark &            &            &            &            &            & 0.222 & 0.312 & 0.401 \\ 
        &          & \checkmark & \checkmark & \checkmark &            &            &            &            & 0.224 & 0.280 & 0.422 \\ 
        &          & \checkmark & \checkmark &            & \checkmark &            &            &            & 0.326 & 0.433 & 0.566 \\ 
        &          & \checkmark & \checkmark &            & \checkmark & \checkmark &            &            & 0.345 & 0.487 & 0.626 \\ 
        &          & \checkmark & \checkmark &            & \checkmark & \checkmark & \checkmark &            & 0.380 & 0.505 & 0.621 \\ 
        & OSG      & \checkmark & \checkmark &            & \checkmark & \checkmark & \checkmark & \checkmark & \textbf{0.441} & \textbf{0.567} & \textbf{0.685} \\ 
        \bottomrule[1pt]
      \end{tabular}
    }
  \end{center}
\end{table*}

\subsection{Ablation Study}\label{ablation}

In this section, we summarize the ablation study of our method.
First, we conduct the ablation experiments for our OMG model, which has been introduced previously.
Moreover, we also present a single-granularity model, named OSG (Observe Single Granularity) for the ensemble.

\noindent \textbf{Ablation of OMG.}
Table \ref{table_ablation1} shows the ablation study of our OMG.
The baseline is a dual-stream architecture with ResNet50 \cite{he2016deep}, BERT \cite{lu2019vilbert} and InfoNCE loss \cite{van2018representation}.
Only the target image and the global text are used as input for the visual stream and textual stream respectively.
Other optimization strategies are illustrated as follows:
\begin{itemize}
  \item BoT: Pre-training a ReID model BoT \cite{luo2019strong} on Track1 with ResNet50-IBN \cite{pan2018two} as the backbone to replace the original image encoder. 
  Experiments shows that the stronger visual encoder brings 0.01 improvement for MRR.
  \item Context: Taking the context image as the input to extract context features.
  With the help of the context information, all metrics MRR, R@5 and R@10 improve. 
  \item Motion: Taking the motion image as the input to extract motion features.
  The use of the foreground motion image doesn't boost the performance, perhaps beacuse of the too rough appearance information in it.
  Therefore, we discard it in the future version of our framework.
  \item Local: Adding three local texts as the input of the textual stream.
  Finer granularities of textual descriptions improve the performance by a large margin (from 0.218 to 0.241).
  \item Prompt: Adding the prompt text as the input of the textual stream.
  Though its simplicity, this simple color-type prompt boosts the performance significantly.
  \item CLIP: Replacing the text encoder BERT with the textual stream of CLIP \cite{radford2021learning}.
  The improvements can be attributed to its excellent ability of modal alignment.
  \item NLAug: Data augmentation with the backtranslation method for the natural language data.
  We can observe a huge gap between the effect of $NLAug$ for OMG and OSG (in Table \ref{table_ablation2}).
  Specifically, it improves the MMR of OMG by 0.085 (from 0.316 ro 0.401), but only 0.004 for OSG (from 0.208 to 0.212).
  The reason behind the gap is that the multi-granularity architecture makes better use of the augmented data.
  \item IDLoss: Applying the auxiliary ID loss.
  The ID loss helps the image encoder achieve more robust representation ability and improves the retrieval performance.
  \item Swin: Replacing the image encoder ResNet50 with the Swin-Transformer-B \cite{liu2021swin}.
  A stronger architecture of the image encoder further boosts the performance.
  \item MFT: Multi-frame testing, i.e., sampling uniformly 8 frames for the target trajectory to generate the mean features instead of only one frame.
  Experiments proves its effectiveness.
\end{itemize}

\noindent \textbf{Ablation of OSG.} 
We also conduct various experiments on a single granularity model (OSG) with different modules, as illustrated in Table \ref{table_ablation2}. 
It has the same baseline with OMG. 
The "NLAug" denotes the same natural language augmentation in Table \ref{table_ablation1}, which can also improve the performance of OSG slightly.
The terms "Color", "Type" and "ID" indicate that we utilize additional classification heads to classify the color, type and ID of the vehicle. 
As shown in the Table \ref{table_ablation2}, the color and ID supervision gains a relative MRR improvement 
which demonstrates the effectiveness of the supervision for the vehicle properties. 
However, the type of the vehicle can hardly help the OSG perform better. 
We infer this is due to the variety descriptions of vehicle type for the same target, such as "truck", "pickup", "hatchback", etc., 
which may lead to confusion. 
We also replace BERT with the text encoder of CLIP, and replace ResNet50 with Swin-Transformer, which gives a great boost to our model. 
To get the correct and representative motion information from the annotation, we first filter out the bounding box 
if its IoU is greater than 0.9 with the box in the previous frames. 
Then we uniformly sample 16 boxes from the remaining ones.
At last, it is inputed into a GRU \cite{cho2014learning} to embed the motion features in the potential semantic space.
The introduction of motion information enhances the model performance, which indicates the motion information plays an important role in this task.

\begin{figure*}[t]
  \centering
  \includegraphics[width = 0.6\textwidth]{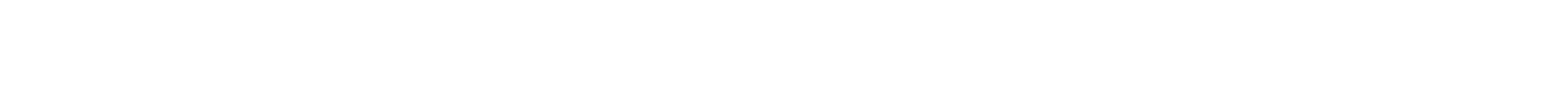}
  \includegraphics[width = 1\textwidth]{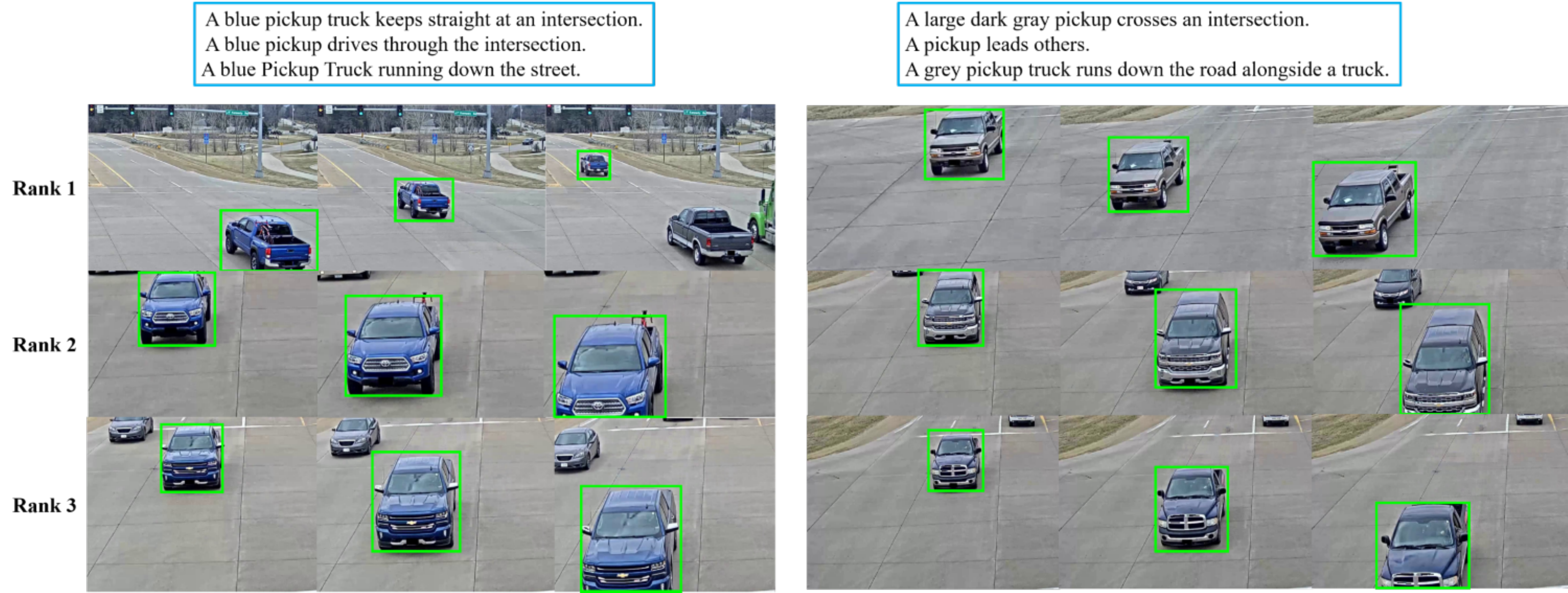}
  \includegraphics[width = 0.6\textwidth]{white.pdf}
  \includegraphics[width = 1\textwidth]{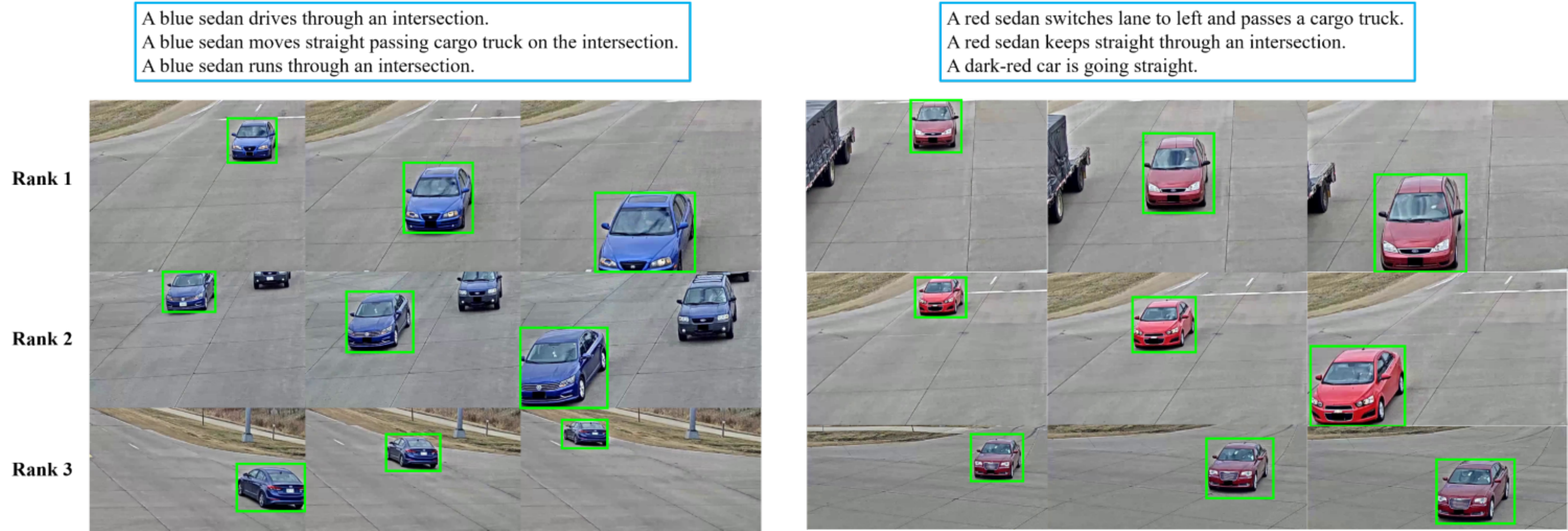}
  \includegraphics[width = 0.6\textwidth]{white.pdf}
  \includegraphics[width = 1\textwidth]{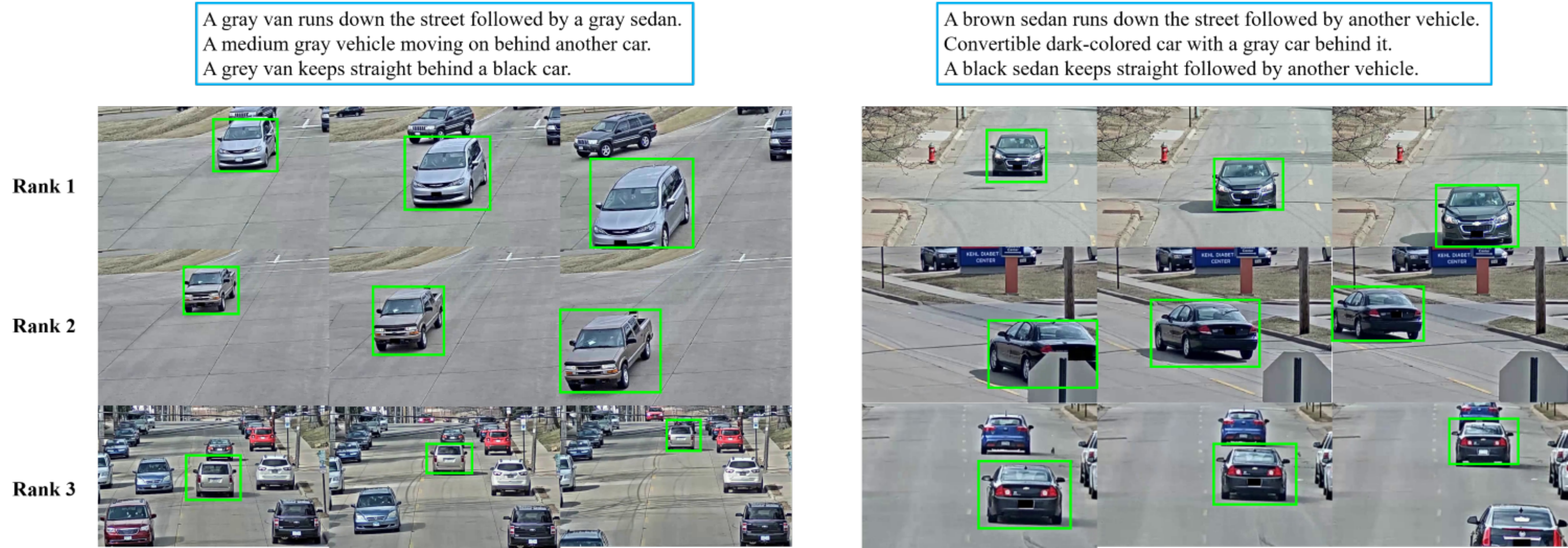}
  \caption{
    Sample retrieval results visualization of our OMG on the test set of the CityFlow-NL dataset.
    Only the rank 1~3 results are listed and only 3 frames are sampled for each trajectory.
    The image is cropped for clarity.
    The target vehicle retrieved are marked with green bounding boxes.
    It's shown that our model has a certain semantic understanding ability.
  }
  \label{figure_example}
\end{figure*}

\subsection{Evaluation Results}

\noindent \textbf{Quantitative Results.} We compare our OMG with previous state-of-the-art methods in Table \ref{table_previous}.
It is shown that our method surpasses them by a large margin.
Specifically, our method is 61.1\% better than the last year's winning team, i.e., 0.3012 vs. 0.1869.
Table \ref{table_leaderboard} lists the top-15 team results on the AI City 2022 Challenge Track2 and we rank the 9th palce.
Note that the final submission are the ensemble results from both OMG and OSG (in Section \ref{ablation})

\noindent \textbf{Qualitative Results.} Figure \ref{figure_example} visualizes the retrieval results of OMG on the test set of the CityFlow-NL dataset.
Only the top-3 results are listed here.

\begin{table}[t]
  \begin{center}
    \caption{
      Comparision with previous methods.
    }
    \label{table_previous}
    \resizebox{0.4\textwidth}{!}{
      \begin{tabular}{cl|c}
        \toprule[1pt]
        & \textbf{Team} & \textbf{MRR} \\
        \hline
        & \textbf{OMG(ours)} & \textbf{0.3012} \\ 
        & Alibaba-UTS-ZJU \cite{1st} & 0.1869 \\ 
        & SDU-XidianU-SDJZU \cite{2nd} & 0.1613 \\ 
        & SUNYKorea \cite{3rd} & 0.1594 \\ 
        & Sun Asterisk \cite{4th} & 0.1571 \\ 
        & HCMUS \cite{6th} & 0.1560 \\ 
        & TUE \cite{7th} & 0.1548 \\ 
        & JHU-UMD \cite{8th} & 0.1364 \\ 
        & Modulabs-Naver-KookminU \cite{10th} & 0.1195 \\ 
        & Unimore \cite{11th} & 0.1078 \\
        \bottomrule[1pt]
      \end{tabular}
    }
  \end{center}
\end{table}

\begin{table}[t]
  \begin{center}
    \caption{
      The leaderboard of the 6th AI City Challenge Track2.
    }
    \label{table_leaderboard}
    \resizebox{0.4\textwidth}{!}{
      \begin{tabular}{cl|c|c}
        \toprule[1pt]
        & \textbf{Rank} & \textbf{Team} & \textbf{MRR} \\
        \hline
        & 1 & Must Win & 0.6606 \\ 
        & 2 & Thursday & 0.5251 \\ 
        & 3 & HCMIU-CVIP & 0.4773 \\ 
        & 4 & MegVideo & 0.4392 \\ 
        & 5 & HCMUS & 0.3611 \\ 
        & 6 & P \& L & 0.3338 \\ 
        & 7 & Terminus-AI & 0.3320 \\ 
        & 8 & MARS\_WHU & 0.3205 \\ 
        & 9 & \textbf{BUPT\_MCPRL\_T2(ours)} & \textbf{0.3012} \\ 
        & 10 & folklore & 0.2832 \\
        & 11 & HYFL & 0.2804 \\
        & 12 & alpha & 0.2802 \\
        & 13 & SEEE-HUST & 0.2333 \\
        & 14 & ETRI\_AIA & 0.0389 \\
        & 15 & Pair Lab & 0,0216 \\
        \bottomrule[1pt]
      \end{tabular}
    }
  \end{center}
\end{table}

\subsection{Limitations}

Our OMG still has several limitations.
First of all, though it can mine rich information from multiple granularities, we simply average the similarities between different cross-modal pairs.
Improving the mechanism for fusing these multi-granularity features is expected to bring significant improvements,
e.g., using the self-attention mechanism \cite{vaswani2017attention}.
Besides, the leverage of the motion and relation information should be further improved.
The temporal information is also underutilized.
Data augmentation is vital for representation learning, which should be studied further.
Last but not least, the cross-modal alignment problem is not trivial for the text-vision retrieval task,
which is not given enough attention in this work.
A more effective architecture or alignment mechanism would help boost the retrieval performance.

\section{Conclusion}

In this paper, we propose a novel framework for the natural language-based vehicle retrieval task, named OMG.
It follows the dual-stream paradigm, which consists of a multi-granularity visual stream and a multi-granularity textual stream.
By focusing on three-granularity of visual information (target, context, motion) 
and three-granularity of textual information (global, local, prompt), 
our model can mine rich semantics jointly with a multi-granularity InfoNCE loss and an auxiliary ID loss.
With some optimization tricks, our method improves the performance of the baseline by 144\% on the validation set.
On the test set, OMG outperforms all previous methods by a large margin and ranks 9th on the 6th AI City Challenge Track2.

In the future, we will focus on discriminative representation learning and cross-modal alignment.

\section{Acknowledgements}
   This work is supported by Chinese National Natural Science Foundation under Grants (62076033, U1931202) 
   and MoE-CMCC "Artifical Intelligence" Project No.MCM20190701.

{\small
\bibliographystyle{ieee_fullname}
\bibliography{egbib}
}

\end{document}